\documentclass[conference,a4paper]{IEEEtran}

\usepackage{fullpage}
\usepackage{amsmath}
\usepackage{graphicx}
\usepackage{times}
\usepackage{epsfig}
\usepackage{graphicx}
\usepackage{amsmath}
\usepackage{amssymb}
\usepackage{supertabular}
\usepackage{graphicx,subfigure}
\usepackage{multirow}
\usepackage[bottom]{footmisc}
\usepackage{url}
\usepackage{color}
\usepackage[T1]{fontenc}
\usepackage[utf8]{inputenc}
\usepackage{authblk}
\usepackage{graphicx}
\usepackage{caption}
\usepackage{epsfig}
\usepackage{amssymb}
\usepackage{amsmath}
\usepackage{amsfonts}
\usepackage{graphicx}
\usepackage{graphicx,subfigure}
\usepackage{url}
\usepackage{subfloat}
\usepackage{times}
\usepackage{balance}
\usepackage[dvipsnames]{xcolor}
\usepackage{algorithm}
\usepackage[noend]{algpseudocode}
\usepackage[
top    = 0.75 in,
bottom = 2in,
left   = 1in,
right  = 1in]{geometry}

\ifCLASSINFOpdf
\else
\fi
\newcommand{\steve}[3]{{\colod{green}}}

\newcommand{\etc}{etc.~}

\newcommand{\footlabel}[2]{%
    \addtocounter{footnote}{1}%
    \footnotetext[\thefootnote]{%
        \addtocounter{footnote}{-1}%
        \refstepcounter{footnote}\label{#1}%
        #2%
    }%
    $^{\ref{#1}}$%
}

\usepackage[pagebackref=true,breaklinks=true,letterpaper=true,colorlinks,bookmarks=false]{hyperref}

\begin{document}
%some lame title for now
\title{Item Popularity Prediction in E-commerce Using Image Quality Feature Vectors}


 \author{
Stephen Zakrewsky\\
\emph{Drexel University}\\
\emph{sz372@drexel.edu}
\and
Kamelia Aryafar\\ 
\emph {Etsy}\\
\emph {karyafar@etsy.com}
\and
Ali Shokoufandeh\\
\emph{Drexel University}\\
\emph{ashokouf@cs.drexel.edu}}

\maketitle

\begin{abstract}
%\boldmath
Online retail is a visual experience- Shoppers often use images as first
 order information to decide if an item matches their personal style. 
 Image characteristics such as color, simplicity, scene composition, texture, style, 
 aesthetics and overall quality play a crucial role in
  making a purchase decision, clicking on or liking a product listing. In this paper we use a set of image features that indicate quality
   to predict product listing popularity on a major e-commerce website, Etsy~\footlabel{etsyurl}{\url{www.etsy.com}}.
   We first define listing popularity through search clicks, favoriting and purchase activity. Next, we infer listing quality from the pixel-level information of listed images as quality features. We then compare our findings to text-only models for popularity prediction. Our initial results indicate that a combined image and text modeling of product listings outperforms text-only models in popularity prediction.
\end{abstract}

\ifCLASSOPTIONpeerreview
 \begin{center} \bfseries EDICS Category: 3-BBND \end{center}
 \fi
\IEEEpeerreviewmaketitle

\section{Introduction}
The informative presentation of product listings through text and 
images is the foundation of modern e-commerce. Shoppers often have a 
specific style or visual preference for many of the available items such as jewelry, clothing, home decor, etc. 
Images provide the first order information for product listings. 
Users often use images in combination with other data modalities such as textual description, price, ratings and \etc to decide if an item is a suitable match for what they need and have in mind. The selection of proper high quality images is then an important step in listing a successful product. In this paper we examine the role of image quality in listing popularity on a major e-commerce website, Etsy~\footref{etsyurl}.

\begin{figure}
  \centering
    \includegraphics[scale=0.15]{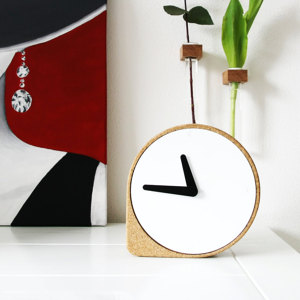}
    \includegraphics[scale=0.15]{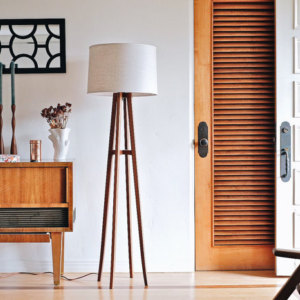}
     \includegraphics[scale=0.15]{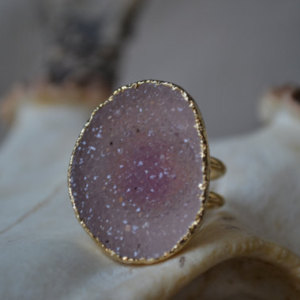}
    \includegraphics[scale=0.15]{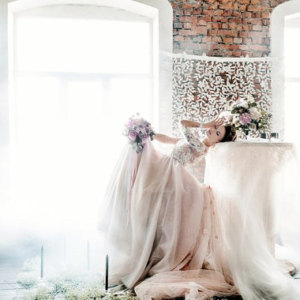}
    \includegraphics[scale=0.15]{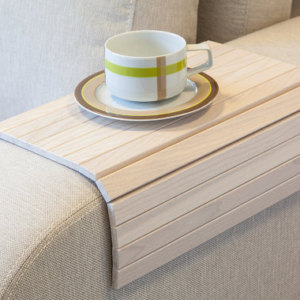}
    \includegraphics[scale=0.15]{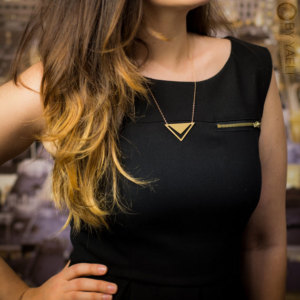}
     \includegraphics[scale=0.15]{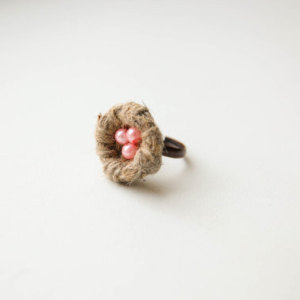}
    \includegraphics[scale=0.15]{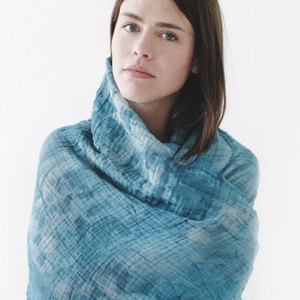}
     \includegraphics[scale=0.15]{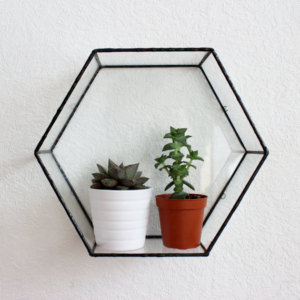}
    \includegraphics[scale=0.15]{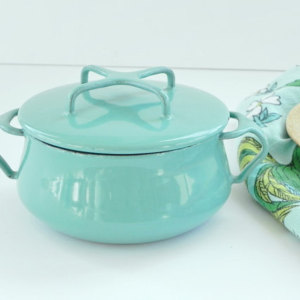}
     \includegraphics[scale=0.15]{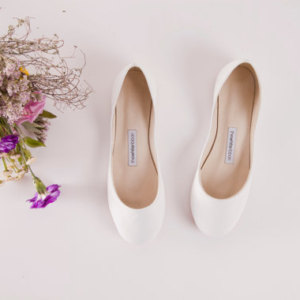}
    \includegraphics[scale=0.15]{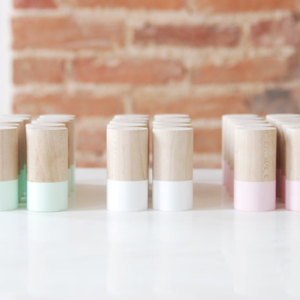}
     \includegraphics[scale=0.15]{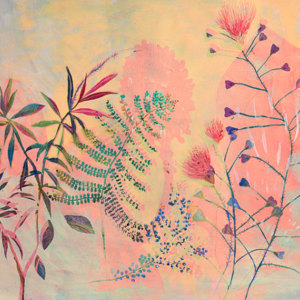}
     \includegraphics[scale=0.15]{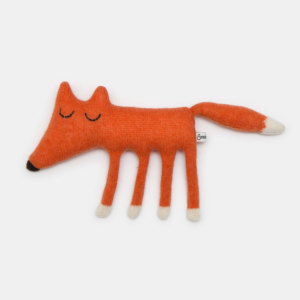}
    \includegraphics[scale=0.15]{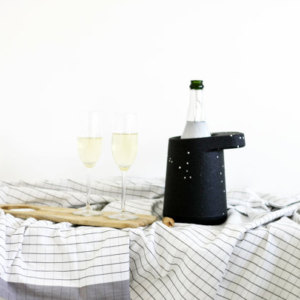}
     \includegraphics[scale=0.15]{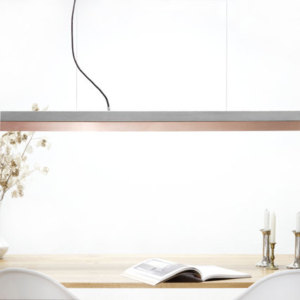}
       \includegraphics[scale=0.15]{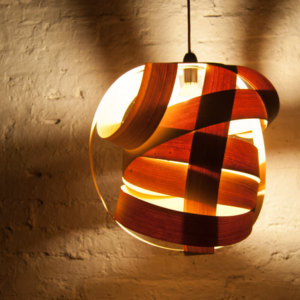}
     \includegraphics[scale=0.15]{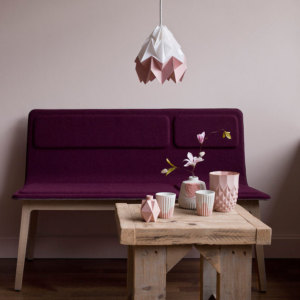}
    \includegraphics[scale=0.15]{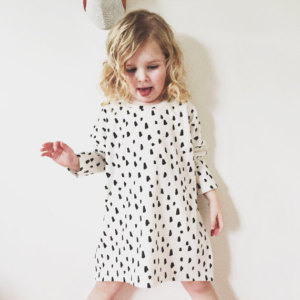}
     \includegraphics[scale=0.15]{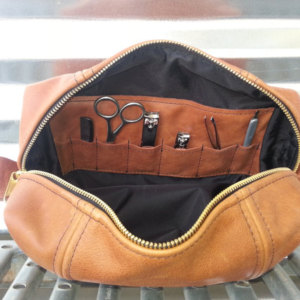}
  \caption{
  Sample Etsy listing images are shown with different lighting, scene composition, and quality.
  }
  \label{fig:etsylistings}
\end{figure}

Etsy is an online marketplace for artisans selling unique handcrafted goods, and vintage wares that
couldn't be found elsewhere. Etsy caters to the long tail of online retail~\cite{Anderson:2006,aryafar2014exploring}. With more than one million sellers, $35$ million unique product listings and nearly a hundred million images, Etsy is uniquely positioned to answer some interesting questions about the role of images as a rich visual experience in e-commerce settings. Each Etsy listing is composed of text information such as title, tags, item description, shop and seller name and complementary images. For a product listing to stand out, high-quality images describing the content of the product listing is a necessity~\cite{wang2011aesthetics,obrador2009role}. Figure~\ref{fig:etsylistings} illustrates some Etsy images with different scene composition, lighting and image quality as featured on the website.

Early work in the literature has defined image popularity as quality~\cite{ke2006design} or aesthetics~\cite{datta2006studying} and use data from photography rating websites where users who have interest in photography upload their photos and rate others. Popularity has also been defined as memorability~\cite{isola2011makes}, and interestingness~\cite{dhar2011high,gygli2013interestingness}.  More recent work has directly tackled popularity.  In \cite{khosla2014makes}, popularity is defined as the number of views on Flickr, and \cite{aryafar2014exploring} uses favorited listings on Etsy.

In this paper we introduce a mechanism for product listings popularity prediction from the images representing those listings. We then explore the correlation between image quality and user interaction with what is for sale. Because sales are rare in comparison to the number of items available on a large site such as Etsy, we look into a combination of mechanisms for interaction, including the number of favorites, purchases and clicks on items to define item popularity. Favorites indicate an interest in an item and are similar to liking mechanisms on other websites such as thumbs-up on Facebook.

Popularity tends to be predicted using typical classifiers such as SVMs or regression \cite{datta2006studying} \cite{khosla2014makes} \cite{chen2014aesthetic} \cite{wang2015automatic}.  Datta et. al. \cite{datta2006studying} uses a two class SVM classifier with a forward selection algorithm to find suitable feature vectors indicating popularity.  By using elastic net to rank feature relevance to aesthetics, and a best first algorithm to find feature sets that minimize the RMSE cross validation error, \cite{wang2015automatic} are able to achieve a 30.1\% improvement compared to \cite{chen2014aesthetic}.  A few have explored other machine learning techniques.  In \cite{ke2006design} a naive Bayes classifier is used and Aryafar et. al \cite{aryafar2014exploring} studied the significance of color in favorited listings on Etsy using logistic regression, perceptron, passive aggressive and margin infused relaxed algorithms.

The features used in popularity prediction model the same qualities professional photographers use such as light, color, rule of thirds, texture, smoothness, blurriness, depth of field, and scene composition \cite{ke2006design} \cite{datta2006studying} \cite{chen2014aesthetic} \cite{wang2015automatic}.  Most of these features are unsupervised, but some such as the spatial edge distribution and color distribution features of \cite{ke2006design} require all of the labeled training data.  Some recent work has looked at semantic object features.  \cite{khosla2014makes} used the popular CNN ImageNet to detect the presence of 1000 difference object categories in the image.  The presence/absence of these categories is used as the feature. In this paper we propose a combination of simplicity, blur, depth of field, rule of thirds and texture features as the image quality representation. We also combine the image representation with text features as a multimodal embedding of items for sale. State-of-the-art studies have often shown that multimodal embeddings of items can outperform single modality representations for multiple prediction, ranking and classification problems~\cite{lynch2015images} \cite{yu2014click} \cite{yu2015learning}.

The remainder of the paper is organized as follows:
Section~\ref{sec:features} describes the image quality feature vectors.  We examine the
performance of image quality features in predicting listing popularity in
section~\ref{sec:experiments}. Finally, we conclude this paper in
section~\ref{sec:conclusion} and propose future research directions.

\section{Features}
\label{sec:features}
The quality features extracted from images are composed of a set of hand-crafted features including simplicity, blur, depth of field, rule of thirds, experimental and texture features. In this section, we explain the details of each subset of features. The implementation of this features is made publicly available~\footnote{We make our feature extraction pipeline for image quality features available at:\\ \url{https://github.com/szakrewsky/quality-feature-extraction}}. The final image quality feature vector is a concatenation of these features. Table~\ref{tab:dimensions} shows the dimensionality of each feature. The dimensionality of the final quality feature vector (image representation) is the sum of all these features. 

  \subsection{Simplicity}
  High quality photos are typically simpler than others.  They often have one subject placed deliberately in the frame.  Sometimes the background is out of focus to emphasize the subject.  Poor quality photographs tend to have cluttered backgrounds and it may be difficult to distinguish the subject of the scene.  We used the four measures of simplicity from \cite{ke2006design}, spatial edge distribution, hue count, contrast and lightness, and blur.

  \subsubsection{Spatial Edge Distribution}
\begin{figure*}
  \centering
  \subfigure[]{
    \includegraphics[scale=0.25]{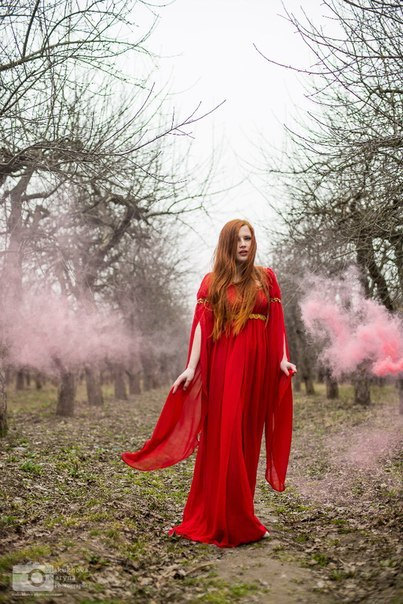}
    \hspace{4.1mm}
    \includegraphics[height=0.235\textheight]{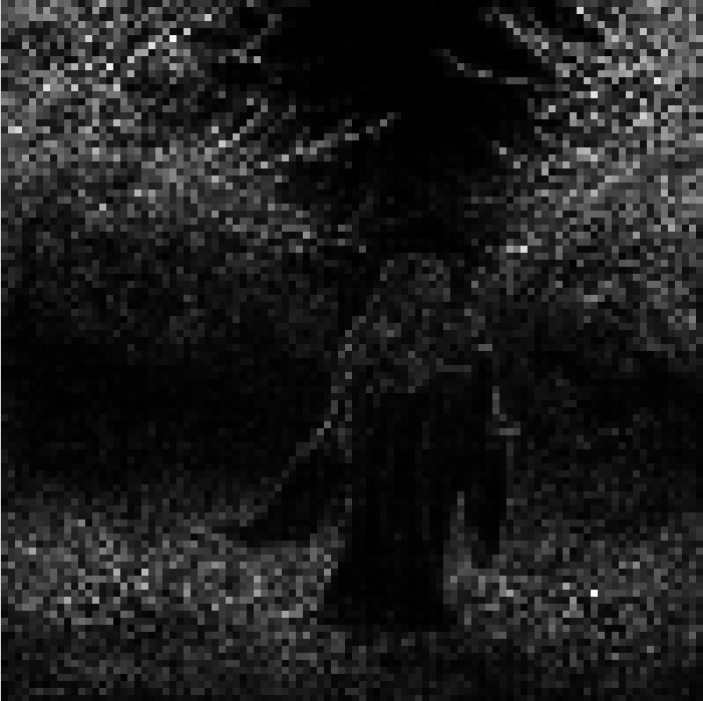}
  }
  \subfigure[]{
    \includegraphics[scale=0.26]{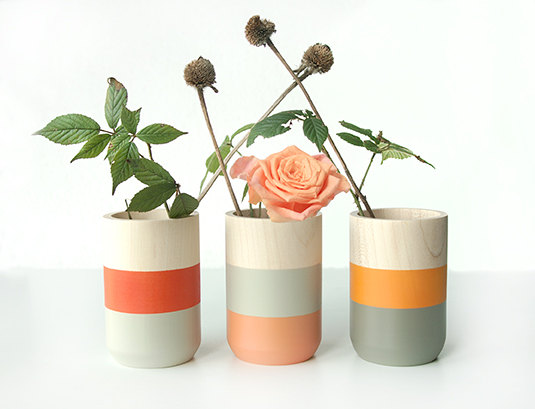}
    \hspace{7mm}
    \includegraphics[scale=0.152]{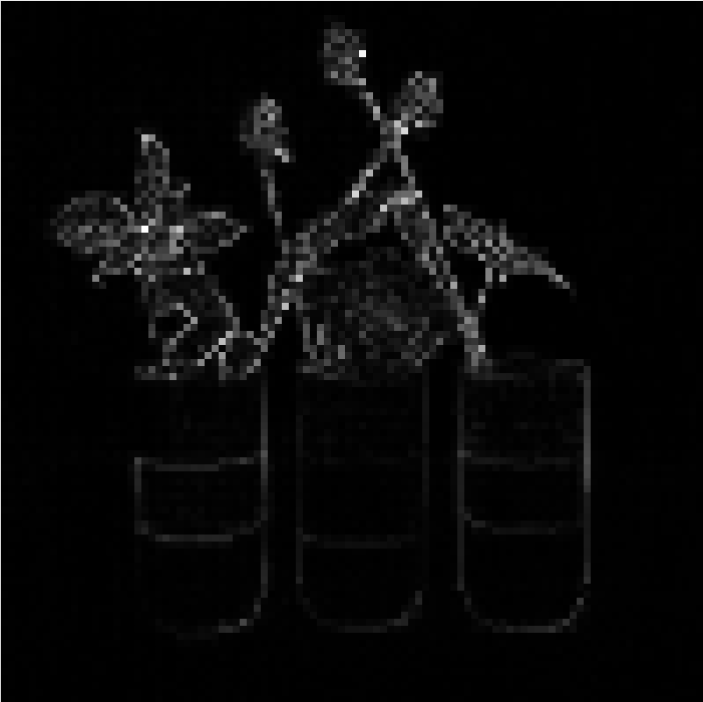}
  }
  \caption{
  The Laplacian image for computing spatial edge distribution for two images is illustrated.  The value of the feature for figure a. is $0.013$ and for b. is $0.30$.
  }
  \label{fig:sed}
\end{figure*}

  Spatial edge distribution measures how spread out sharp edges are in the image.  A single subject is expected to have a small distribution while an image with a cluttered background would have a large distribution.  Edges are detected by applying a $3\times3$ Laplacian filter and taking the absolute value.  The filter is applied to each RGB channel independently and the final image is computed as the mean across all three channels.  The Laplacian image is resized to $100\times100$ and normalized to sum to 1.  Then, the edges are projected onto the $x$ and $y$ axis independently.  Let $w_x$, and $w_y$ be the width of $98\%$ of the projected edges respectively.  The image quality feature $f = 1 - {w_x w_y \over 100}$ is the percent of area outside the majority of edges.  Figure \ref{fig:sed} shows the edges detected from two different images and their respective feature values.

  \subsubsection{Hue Count}
  Professional photographs look more colorful and vibrant, but actually tend to have less distinct hues because cluttered scenes contain many heterogeneous objects.  We use a hue count feature by filtering an image in the HSV color space such that V is in the range of $[0.15, 0.95]$ and S is greater than $0.2$.  A $20$ bin histogram is computed on the remaining H values.  Let $m$ be the maximum value of the histogram and let $N = \{i | H(i) > \alpha m\}$, be the set of bins with values greater than $\alpha m$.  The quality feature $f = 20 - ||N||$ is $0$ when there are a many different hues and grows larger as the number of distinct hues in the image goes down.  We used $alpha = 0.05$ as shown in the literature \cite{ke2006design}.

  \subsubsection{Contrast and Lightness}
  Brightness is a well known variable that professional photographers are trained to understand and adjust.  We use the average brightness feature \cite{ke2006design}, \cite{chen2014aesthetic} computed from the L channel of the Lab color space.  Contrast is similar, and is the ratio of maximum and minimum pixel intensities.  We sum the RGB level histograms, and normalize it to sum to 1.  We use the width of the center $98\%$ mass of the histogram \cite{ke2006design}.

   \subsection{Blur}
  Blurry images are almost always considered to be of poor quality.  We use the common blur features in the literature \cite{ke2006design} \cite{tong2004blur}.  In \cite{ke2006design} blur is modeled as $I_b = G_\sigma * I$ where $I_b$ is the result of convolving a Gaussian filter with an image.  The larger the $\sigma$ the more high frequencies are removed from the image.  Assuming the frequency distribution of all $I$ is approximately the same, then the maximum frequency $||C||$ can be estimated as $C = \{(u, v)\ |\ ||FFT(I_b)|| > \Theta\}$.  The feature is $f = ||C|| \sim 1/\sigma$, after normalizing by the image size.

  In \cite{tong2004blur}, blur estimation is done based on changes in the edge structures.  The blur operation will cause gradual edges to lose sharpness.  Assuming that most images have gradual edges that are sharp enough, the blur is measured as the ratio of gradual edges that have lost their sharpness.
\begin{figure*}[t!]
  \centering
  \subfigure[]{
    \includegraphics[width=0.3\textwidth]{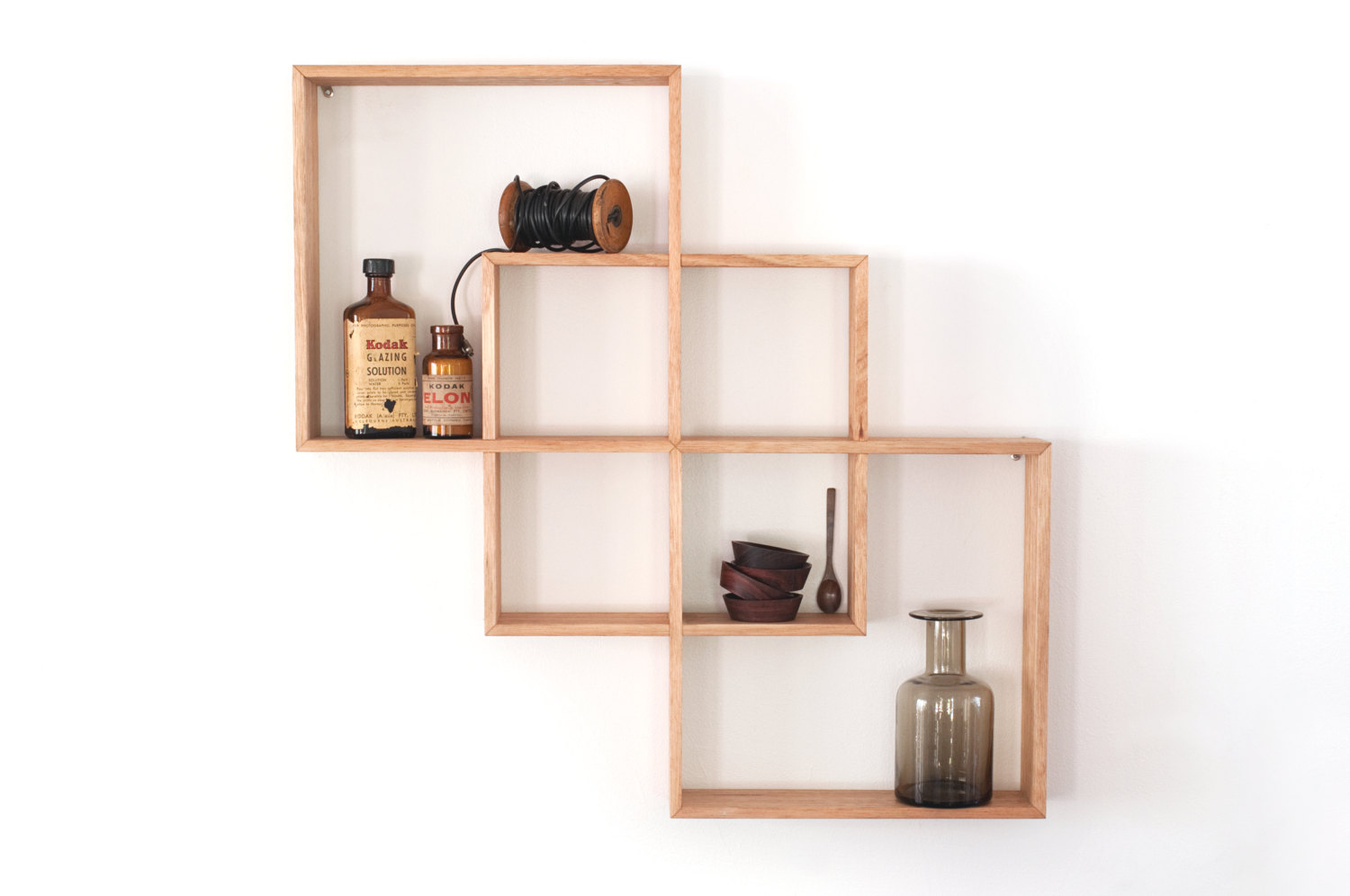}
  }
  \subfigure[]{
    \includegraphics[width=0.3\textwidth]{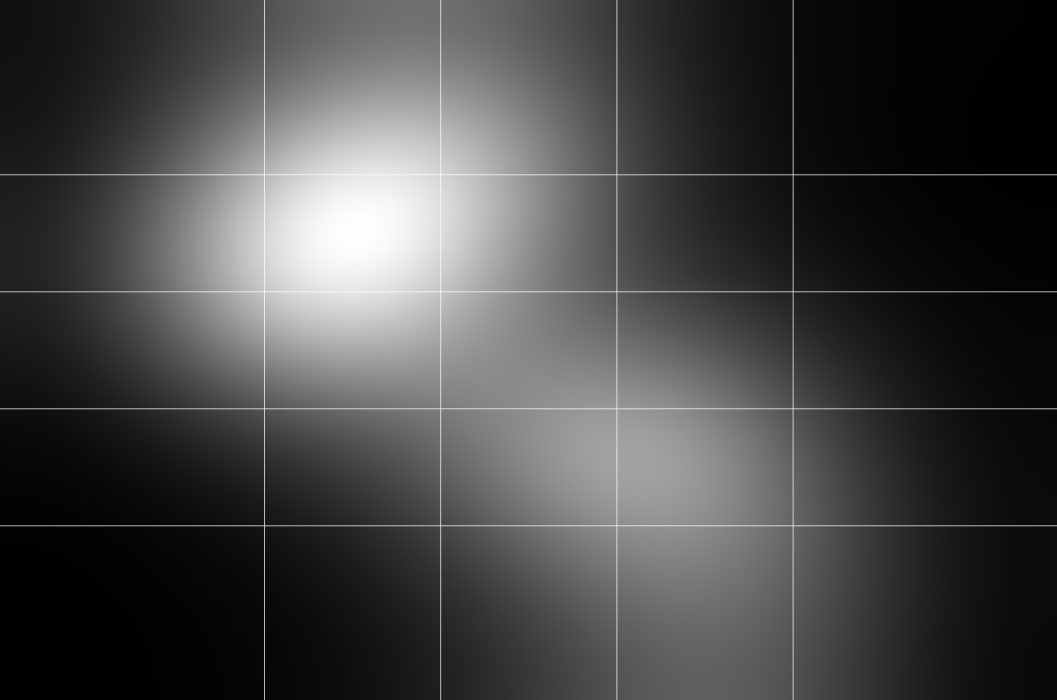}
  }
  \subfigure[]{
    \includegraphics[width=0.2\textwidth]{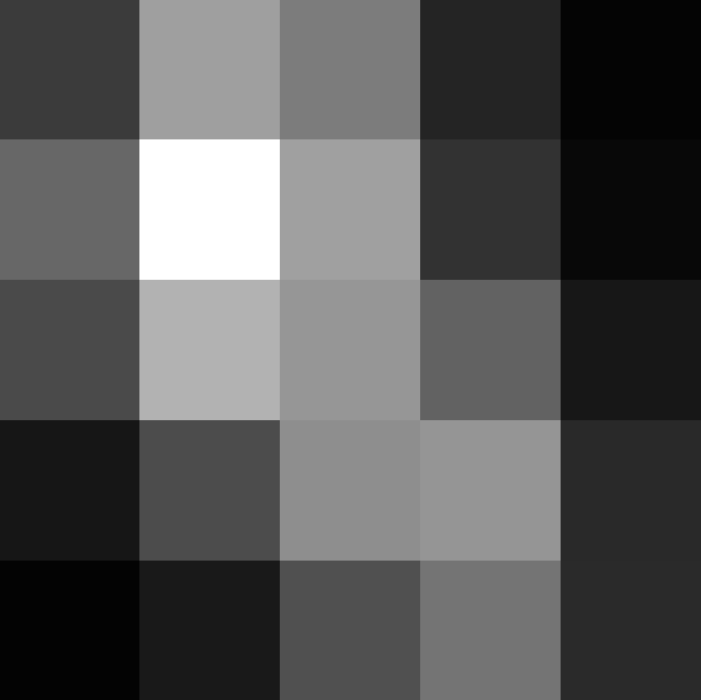}
  }
  \caption{
    Example of Rule of Thirds feature.  Figure b. shows the SR saliency detection, and c. shows the thirds map feature.
  }
  \label{fig:rot}
\end{figure*}

  \subsection{Rule of Thirds}
  The rule of thirds is an important composition technique.  Thirds lines are the horizontal and vertical lines that divide an image into a $3\times3$ grid of equal sized cells.  The rule of thirds states that subjects placed along these lines are aesthetically more pleasing and more natural than subjects centered in the photograph.  In order to segment the subject of the image from the background, we use the Spectral Residual saliency detection algorithm \cite{hou2007saliency}.  The feature is a $5\times5$ map where each cell is the average saliency value \cite{mai2011rule}.  Let $w_p$ be the saliency value of the pixel and $A(W_i)$ is the area of the cell, then the value of each cell is
  \begin{equation}
    w_i = {\sum_{p \in W_i} w_p \over A(W_i)} .
  \end{equation}
  To compute the feature, the image is divided into a $5\times5$ grid with emphasis on the thirds lines; the horizontal and vertical regions centered on the thirds lines are $1/6$ of the image size.  Figure \ref{fig:rot} shows the saliency detection with the $5\times5$ grid overlay, and the thirds map feature for an image.

  \subsection{Texture}
  A smooth image may indicate blur or out-of-focus, and the lack of which may indicate poor film, or too high an ISO setting.  In contrast, texture in the scene is an important composition skill of a photographer.  Smoothness may indicate the lack of texture.  Texture and smoothness are some of the most statically correlated features for quality/popularity \cite{wang2015automatic} \cite{khosla2014makes}.  We use three smoothness/texture features from these.

  A three level wavelet transform is applied to the L channel of the Lab color space.  We only use the bottom level of the pyramid.  The result is squared to indicate power.  Let $b = \{HH, HL, LH\}$ be the bottom level of a wavelet transform, the extracted feature is then
  \begin{equation}
    f = {1\over 3MN} \sum_{m=1}^{M}\sum_{n=1}^{N}\sum_{b}w^b(m, n)
  \end{equation}
  where $w$ is the square of the wavelet value.  Because the Laplacian is often used as a pyramid of different scales, another feature
  \begin{equation}
    f = {1\over MN} \sum_{m=1}^{M}\sum_{n=1}^{N}l(m, n)
  \end{equation} is also used.  This time $l$ is the second level from the bottom of a Laplacian pyramid.

\begin{figure*}
  \centering
  \subfigure[]{
    \includegraphics[width=0.3\textwidth]{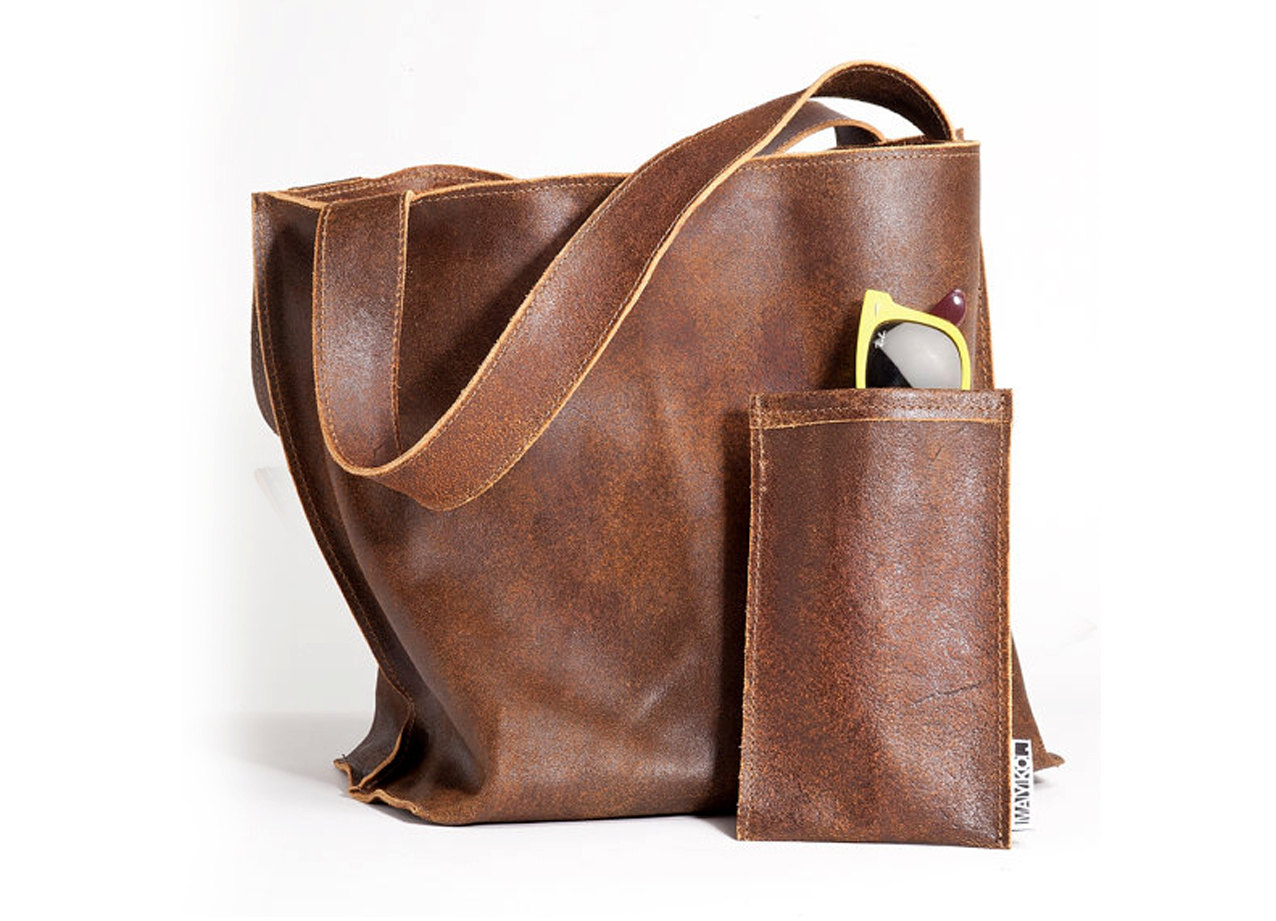}
  }
  \subfigure[]{
    \includegraphics[width=0.3\textwidth]{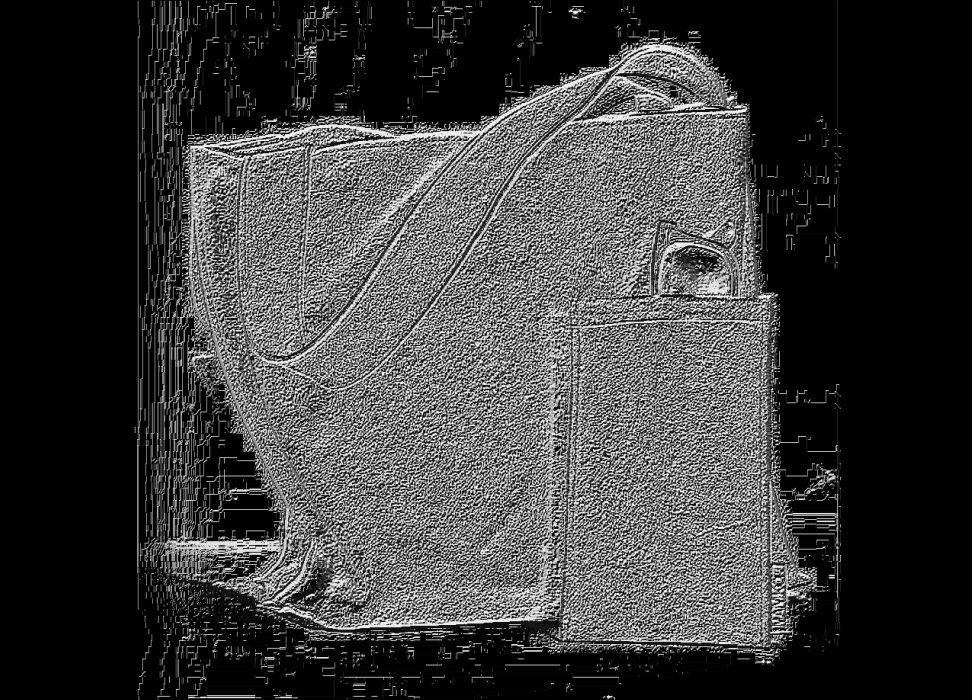}
  }
  \subfigure[]{
    \includegraphics[width=0.3\textwidth]{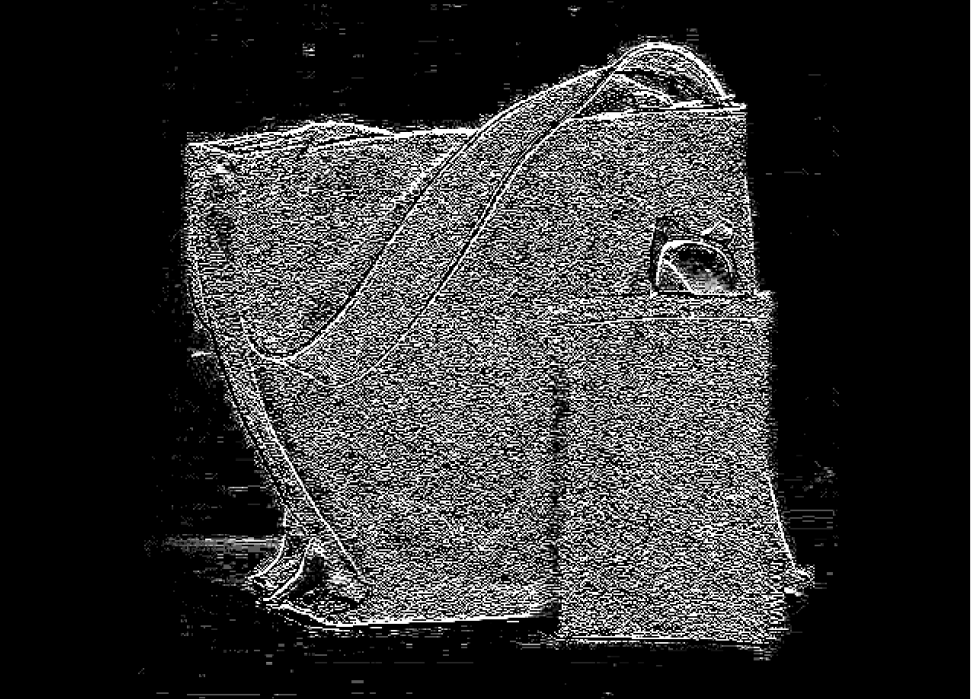}
    \includegraphics[width=0.3\textwidth]{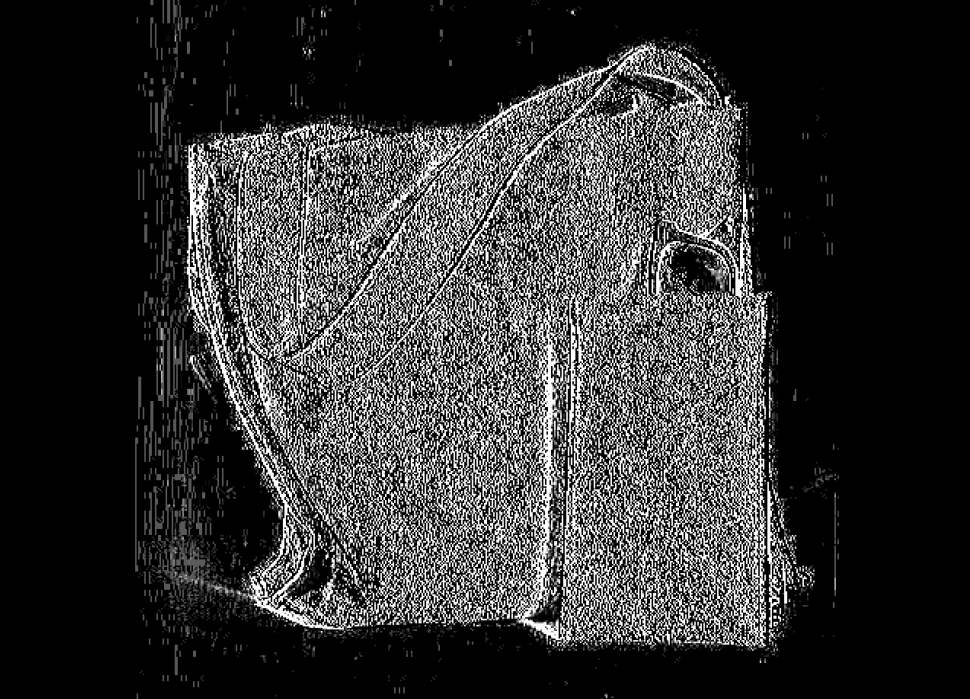}
    \includegraphics[width=0.3\textwidth]{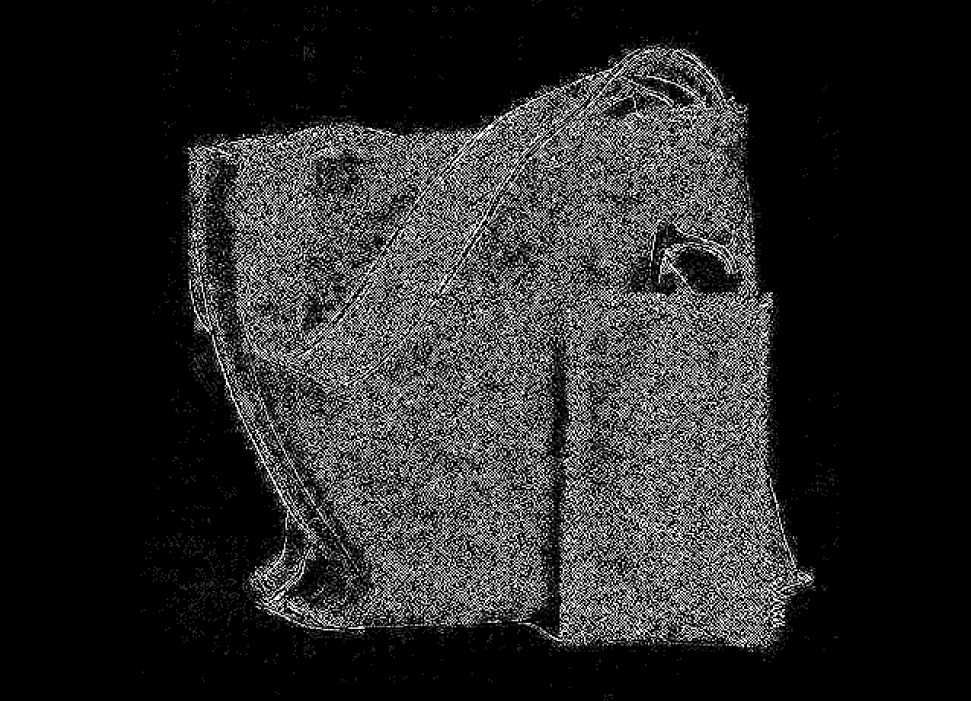}
  }
  \caption{
  Smoothness and texture features are illustrated.  Figure b. shows Local Binary Pattern (LBP) feature image, and c. shows the 3 channels of the DB1 wavelet transform on the sample image.
  }
  \label{fig:tex}
\end{figure*}

  Another texture feature is computed using local binary pattern (LBP).  Then a pyramid of histograms are computed as in \cite{lazebnik2006beyond}.  Figure \ref{fig:tex} shows the similarities of LBP features and the three channels of Daubechies db1 wavelet.

  \subsection{Depth of Field}
\begin{figure*}[t!]
  \centering
  \subfigure[]{
    \includegraphics[width=0.3\textwidth]{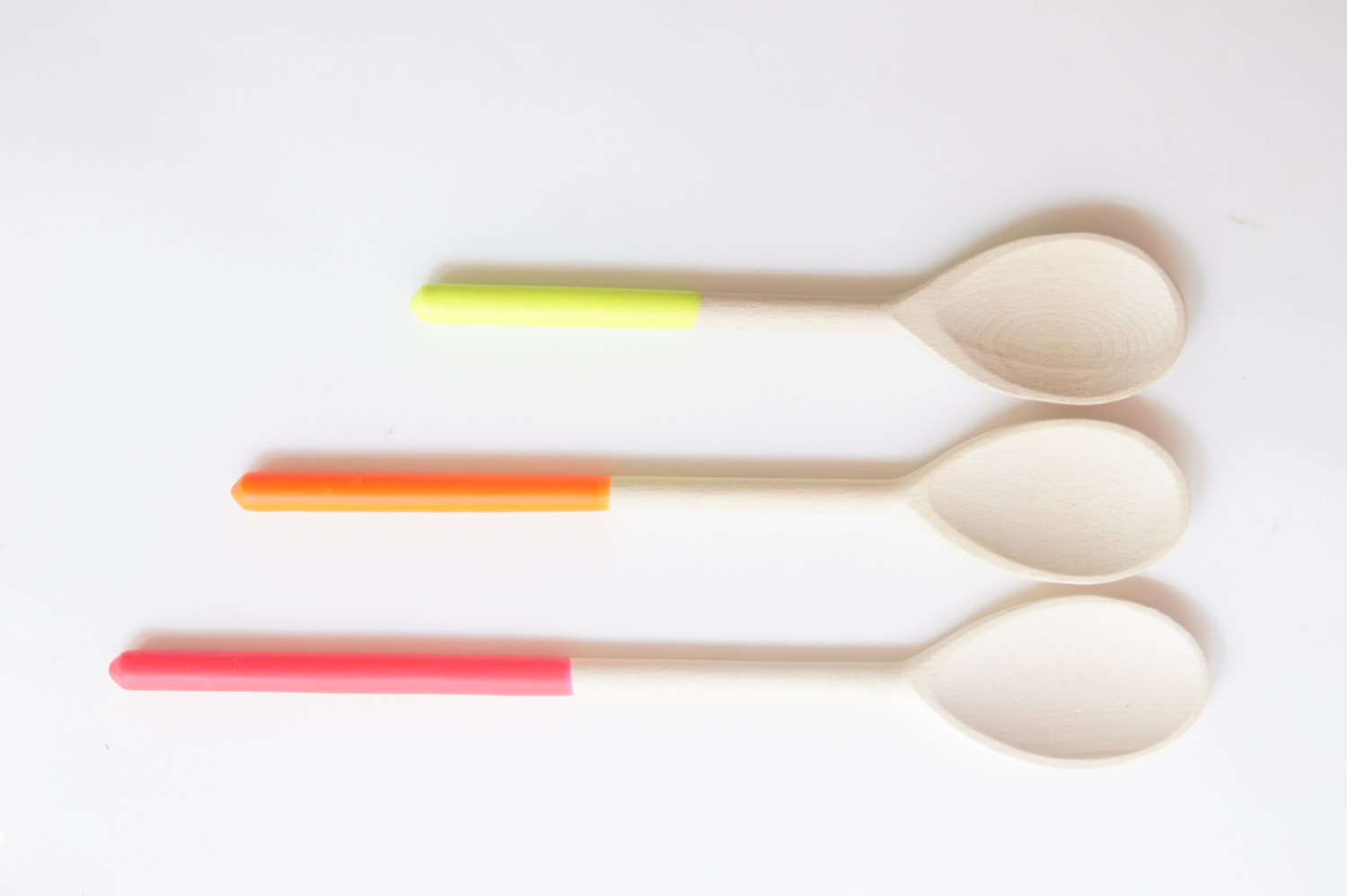}
  }
  \subfigure[]{
    \includegraphics[width=0.304\textwidth]{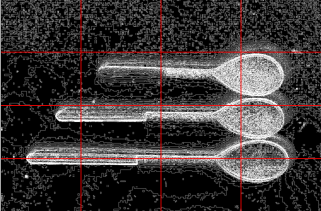}
  }
  \subfigure[]{
    \includegraphics[width=0.3\textwidth]{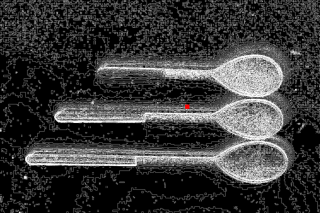}
  }
  \caption{
  Figure b. shows the Low Depth of Field features in the center grid region for the Laplacian image.  Figure c. shows the same image with its center of mass.
  }
  \label{fig:lowdof}
\end{figure*}

  Depth of field is the distance between the nearest and farthest objects that appear in sharp focus.  A technique of professional photographers is to use low depth of field to focus on the photographic subject while blurring the background.  We used the feature \cite{datta2006studying} of the ratio of high frequency detail in center regions of the image compared to the entire image.  Let $w$ be the bottom level of a wavelet transform, the feature can be describes as:
  \begin{equation}
    f ={\sum_{(x,y)}\in M_6\cup M_7\cup M_{10}\cup M_{11} w(x,y) \over \sum_{i=1}^{16}\sum_{(x,y)\in M_i} w(x,y)} ,
  \end{equation}
  where $M_i | 1 \le i \le 16$ are the cells of a $4\times4$ grid.  The same feature is also reapplied using the Laplacian pyramid $l$ instead of $w$ \cite{wang2015automatic}.  These features only look at the center region of the image.  A third feature \cite{wang2015automatic} looks at the spatial distribution of high frequency details.  Let $l$ be the bottom layer of a Laplacian pyramid and $c_{row}, c_{col}$ are the center of mass, the feature is obtained as:
  \begin{equation} 
    f = {1 \over MN} \sum_{m=1}^M\sum_{n=1}^N l(m,n)\sqrt{(m-c_{row})^2 + (n-c_{col})^2} .
  \end{equation}
  Figure \ref{fig:lowdof} visualizes how these features are computed for a sample image.

  \subsection{Experimental}
  Maximally Stable Extremal Regions (MSER) \cite{matas2004robust} can be used to detect text because characters are typically single solid colors with sharp edges that standout from the background \cite{chen2011robust}.  Additionally, texture patterns are also often detected by MSER, like bricks on a wall.  In this paper, we used the count of the number of MSER regions as the experimental feature. In the future, we would like to continue this experiment into other features based on text in images.
 
\section{Popularity Prediction}
\label{sec:experiments}
We collect a set of images from Etsy through Etsy's API\footnote{\url{www.etsy.com/developers}} for popularity prediction.
Our dataset consists of $50,000$ Etsy listing images. Each Etsy listing has at least one photo and can have up to five photos to show different angles and details. In our experiments we only extract the first (main) listing image which shows up in search results and is featured as the main image on the listing page. We denote the number of favorites for each listing, $L$ with main image $I$ as $F(L_I)$, the number of purchases with $P(L_I)$ and number of clicks with $C(L_I)$. We associate each listing image with it's popularity score as :
$$Popularity(L_I) = \sum F(L_I) + C(L_I) + P(L_I).$$

\begin{table}
\label{tab:dimensions}
\begin{center}
\begin{tabular}{| l | c | c |}
\hline
Feature & Dimension \\ \hline
'Ke06-qa': spatial edge distribution & 1 \\ \hline
'Ke06-qh': hue count & 1 \\ \hline
'Ke06-qf': blur & 1 \\ \hline
'Ke06-tong': blur tong etal & 1 \\ \hline
'Ke06-qct': contrast & 1 \\ \hline
'Ke06-qb': brightness & 1 \\ \hline
'-mser count': mser count & 1 \\ \hline
'Mai11-thirds map': thirds map & 25 \\ \hline
'Wang15-f1': avg lightness & 1 \\ \hline
'Wang15-f14': wavelet smoothness, & 1 \\ \hline
'Wang15-f18': laplacian smoothness & 1 \\ \hline
'Wang15-f21': wavelet low dof & 1 \\ \hline
'Wang15-f22': laplacian low dof & 1 \\ \hline
'Wang15-f26': laplacian low dof swd & 1 \\ \hline
'Khosla14-texture': texture & 5120 \\ \hline
\hline
\end{tabular}
 \label{tab:dimensions}
\end{center}
\caption{Image quality feature dimensions are shown by feature.}
\label{tab:dimensions}
\end{table}

\begin{table}[h!]
   \label{tab:results}
   \caption{Lift in accuracy rate using a logistic regression, relative to text-only baseline ($\%$), on the sample dataset is shown in image-only and multimodal settings.}
    \begin{center}
\begin{tabular}{|c||c|c|c|} 
\hline
Modality&Image&Image+Text (MM)\\
\hline
 \hline
Relative lift in AUC& $+1.07\%$ &$+\mathbf{3.45}\%$\\ \hline
 \end{tabular}
 \label{tab:results}
   \end{center}
\end{table}

\newcommand{\tuple}[1]{\ensuremath{\left \langle #1 \right \rangle }}

We extract the quality feature vectors as described in Section~\ref{sec:features} for each listing image and denote that with $q(L_I)$ for listing $L$ and image $I$. Table~\ref{tab:dimensions} shows the dimensionality of each feature that is used to build the quality feature vector. Once the dataset has been tagged with these quality features, we extract textual information from the
listing as $t(L_I)$. These textual features consist of the tokenized listings
titles unigrams and bigrams and tokenized listings
tags unigrams and serve as the single modality listing representation. The multimodal feature vector representation, $MM(L_I)$ is obtained by concatenating quality and textual features as a single feature vector, i.e., $MM(L_I) = \tuple{q(L_I),t(L_I)}$. 

We then use a logistic regression against popularity scores, $Popularity(L_I)$ 
and report the accuracy lift using images and multimodal feature vectors relative to the baseline text-only model. Table~\ref{tab:results} shows these results. We can observe that the quality features in combination with textual features can increase the prediction accuracy on the collected dataset.  

\section{conclusion}
\label{sec:conclusion}
This works presents an initial study on understanding how image quality
can impact the popularity of items in e-commerce settings, thereby providing better user
understanding and a better overall shopping experience. To facilitate
this understanding, this work proposed an empirical method to
estimate the image quality features representing product listings
on Etsy. These feature vectors were combined with traditional textual features to serve as the multimodal item representation. We compared the efficiency of single modality (text-only and image-only) features to multimodal feature vectors in popularity prediction. Our initial results indicate that quality features in combination with text information can increase the prediction accuracy for a sample dataset.

\bibliographystyle{IEEEtran}
\bibliography{ref}

\end{document}